\documentclass[10pt,twocolumn,letterpaper]{article}

\usepackage{iccv}              % To produce the CAMERA-READY version
\definecolor{iccvblue}{rgb}{0.21,0.49,0.74}
\usepackage[pagebackref,breaklinks,colorlinks,allcolors=iccvblue]{hyperref}
\usepackage{booktabs} 
\usepackage{multirow}
\usepackage{graphicx}
\usepackage{array}
\usepackage{rotating}
\usepackage{xcolor}    
\usepackage{pifont}
\usepackage[accsupp]{axessibility}

\def\paperID{6842} % *** Enter the Paper ID here
\def\confName{ICCV}
\def\confYear{2025}

\title{DepthSync: Diffusion Guidance-Based Depth Synchronization for \\Scale- and Geometry-Consistent Video Depth Estimation}

\author{
  Yue-Jiang Dong\textsuperscript{1} \quad
  Wang Zhao\textsuperscript{2} \quad
  Jiale Xu\textsuperscript{2} \quad
  Ying Shan\textsuperscript{2} \quad
  Song-Hai Zhang\textsuperscript{1}$^{\ast}$ \\
  \textsuperscript{1}Tsinghua University \quad
  \textsuperscript{2}ARC Lab, Tencent PCG \\
  {\tt \small \url{https://yuejiangdong.github.io/depthsync}}
}

\begin{document}

\twocolumn[{%
\renewcommand\twocolumn[1][]{#1}%
\maketitle
\begin{center}
    \centering
    \vspace{-0.5cm}
    \captionsetup{type=figure}
    \includegraphics[width=\textwidth]{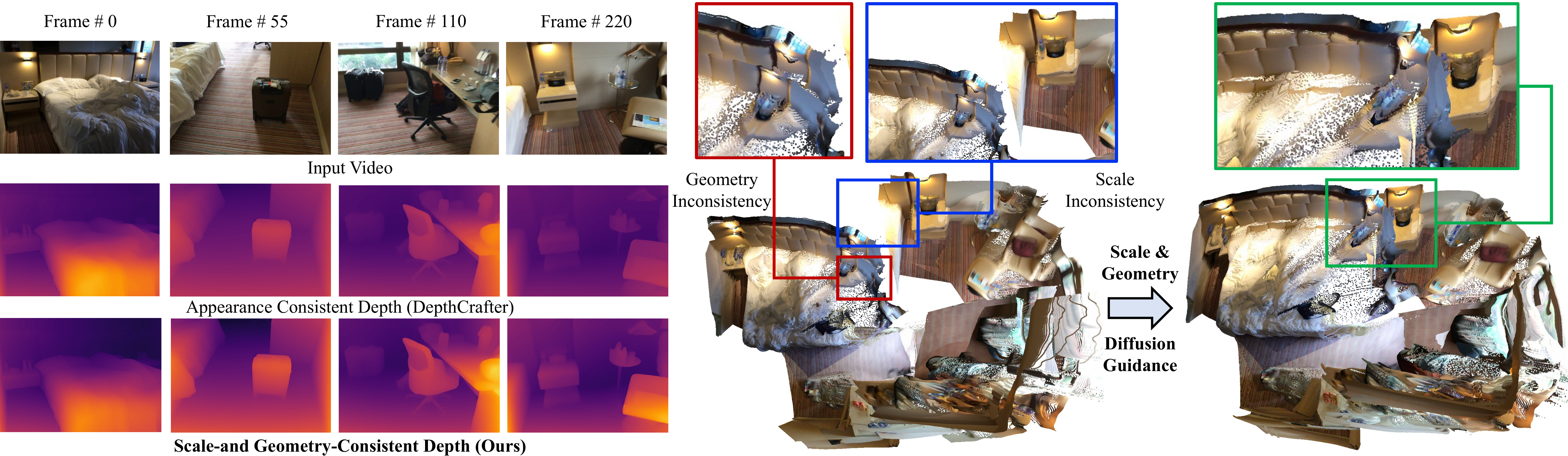} 
    \captionof{figure}{We present \textbf{DepthSync}, a framework that introduces cross-window depth scale synchronization and intra-window geometry alignment for long video depth predictions with enhanced scale and geometry consistency via diffusion guidance. %By combining both strong diffusion priors with the scale and geometric optimization objectives, GD$^{2}$ produces both temporally and geometrically consistent video depth with enhanced accuracy, facilitating high-quality 3D reconstructions. 
    }
    \label{fig:teaser}
\end{center}%
}]

\renewcommand{\thefootnote}{}
\footnotetext[0]{ $^\ast$Corresponding author.}

\maketitle

\begin{abstract}
Diffusion-based video depth estimation methods have achieved remarkable success. % with strong generalization ability. 
However, predicting depth for long videos remains challenging. Existing methods typically split videos into overlapping sliding windows, leading to accumulated scale discrepancies across different windows, particularly as the number of windows increases. Additionally, these methods rely solely on 2D diffusion priors, overlooking the inherent 3D geometric structure of video depths, which results in geometrically inconsistent predictions.
In this paper, we propose DepthSync, a novel, training-free framework using diffusion guidance to achieve scale- and geometry-consistent depth predictions for long videos. Specifically, we introduce \textbf{scale guidance} to synchronize the depth scale \textbf{across windows} and \textbf{geometry guidance} to enforce geometric alignment \textbf{within windows} based on the inherent 3D constraints in video depths. These two terms work synergistically, steering the denoising process toward consistent depth predictions. Experiments on various datasets validate the effectiveness of our method in producing depth estimates with improved scale and geometry consistency, particularly for long videos.
\end{abstract}
\section{Introduction}
Consistent video depth estimation is crucial for accurately perceiving real 3D structures and has broad applications across various fields, including 3D/4D reconstruction \cite{huang20242d, wang2024shape, lei2024mosca}, 3D content creation \cite{long2024wonder3d}, robotics \cite{dong2022towards}, and autonomous driving \cite{zhou2017unsupervised, yurtsever2020survey, dong2024mal}. Recent advancements in leveraging diffusion model priors have demonstrated significant potential for video depth estimation \cite{shao2024chrono, hu2024depthcrafter}. These methods rely on 2D image-to-video diffusion priors \cite{blattmann2023stable, ho2022video}, enabling them to predict temporally consistent depth for videos while achieving enhanced zero-shot generalization capabilities and producing rich depth details.

However, these methods still face limitations in achieving consistent depth predictions across frames, which restricts their applicability in real-world scenarios. One major challenge is maintaining consistency for long input videos. In 2D video diffusion models, consistency primarily relies on cross-attention modules within the diffusion network. However, for long video depth prediction, computational resource limitations necessitate splitting the video into shorter sliding windows, making it difficult to preserve depth consistency across multiple windows. A common solution is to divide the video into overlapping windows and propagate consistency information through the overlapping regions in a window-by-window manner. Previous methods \cite{hu2024depthcrafter, shao2024chrono} inject this information by using predictions from the previous overlapping region to initialize the current window. However, such an approach has limited effectiveness, and the accumulation of scale discrepancies between windows as video length and window count increases leads to severe depth scale misalignment and significant gaps in 3D spatial distance, as illustrated in the blue box depicted in Fig. \ref{fig:teaser}.

On the other hand, while 2D diffusion priors ensure appearance consistency within sliding windows, they lack awareness of inherent geometric alignment between depth frames. As shown in the red box in Fig. \ref{fig:teaser}, the headboards of the bed are misaligned in 3D space across different frames. While this discrepancy is subtle in 2D visualization, it results in significant errors in 3D physical distance. Such geometric inconsistencies are difficult to address using 2D visual priors alone without explicit 3D geometric constraints.

To address these issues, we propose DepthSync, a novel framework for scale- and geometry-consistent depth estimation, particularly for long videos. Our approach adopts the overlapping sliding window scheme for inference, as it is well-suited for real-world applications and supports streaming video input. Unlike prior methods that inject consistency information only at the start of the diffusion process, we pioneer \textbf{scale guidance}, a novel mechanism that enforces scale synchronization at \textit{every step} throughout the denoising trajectory. We leverage diffusion guidance to align the depth of the current window with the previous using least squares optimization on the overlapping region. This alignment guides the denoising process toward a depth scale consistent with the computed aligned result.

Additionally, leveraging the inherent geometric relationships between depth frames, we propose \textbf{geometry guidance} to integrate diffusion, rooted in robust 2D priors, with traditional 3D geometric optimization. This aligns video depths with their inherent geometry through explicit regularization using multiple complementary 3D constraints. The scale and geometry guidance operate sequentially and synergistically, producing scale- and geometry-consistent depth predictions.
% to align the video depths' inherent geometry. We explicitly regularize the denoising process with multiple complementary 3D geometric constraints. The scale and geometry guidance operate sequentially and mutually reinforce each other, resulting in a scale- and geometry-consistent video depth prediction.
% inspired by the notable similarity between the diffusion denoising process and iterative traditional geometry optimization, 

Quantitative and qualitative experimental results demonstrate the effectiveness of our method in improving depth estimation accuracy and consistency across various datasets, particularly for long videos. 
% We further provide extensive evaluation on the poses derived from our predicted depths, validating their geometric plausibility for accurate 3D reconstruction. 

% In summary, our contributions are as follows:
% \begin{itemize}
%     \item We present a novel method for predicting both scale and geometry consistent video depths by injecting regularization along diffusion denoising steps.
%     \item 
%     % \item We propose gradient accumulation and sliding update strategies to accommodate long-term depth generation with enhanced consistency.
%     \item  
%     % \item 
% \end{itemize}

%Classical SLAM and Structure-from-Motion systems~\cite{schonberger2016structure, mur2015orb} mainly rely on optimization algorithms like Bundle Adjustment, upon low-level visual cues like sparse~\cite{lowe2004distinctive, rublee2011orb} or dense pixel matching~\cite{liu2010sift, sun2021loftr, teed2020raft}, to solve global consistent depths in 3D. Despite the effectiveness, they often suffer from the ambiguities in monocular RGB video, such as poorly textured areas, repetitive patterns, challenging lighting conditions, occlusions, and motion blurs \cite{luo2020consistent}, etc., where optimization objectives fail to stay discriminative. 

\begin{figure*}[t!]
    \centering
    \includegraphics[width=\textwidth]{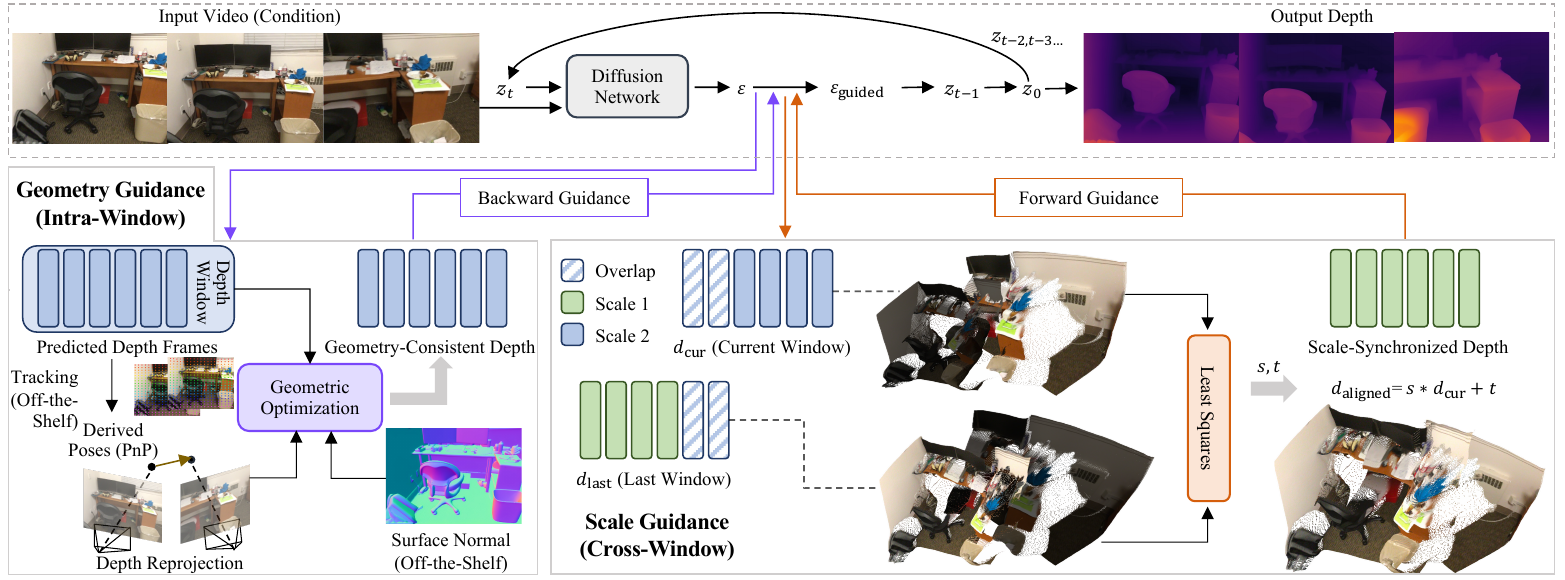}
    \caption{\textbf{Overview.} Our DepthSync framework predicts depth for a monocular video by applying guidance to a pre-trained diffusion-based depth estimation model, ensuring scale- and geometry-consistent depth predictions. Following common practice, the input video is split into overlapping windows and processed sequentially. During denoising, we derive depths from noise prediction, applying geometry guidance to align frames within each window using geometric constraints and scale guidance to synchronize depth scales across windows.}
    \label{fig:overview}
    % \vspace{-10pt}
    \vspace{-0.5cm}
\end{figure*}

\section{Related Work}
\textbf{Monocular Depth Estimation.} Monocular depth estimation employs deep neural networks to regress dense depth maps from images. Research progress has been made in both discriminative \cite{eigen2014depth,fu2018deep,lee2019big,yuan2022neural,bhat2021adabins,bhat2023zoedepth,hu2024metric3d, piccinelli2024unidepth} and generative methods \cite{ji2023ddp, saxena2024surprising} in this field. Notable discriminative methods include MiDaS \cite{ranftl2020towards} and Depth Anything \cite{yang2024depthanything}, which are trained on mixed dataset to enhance generality. Recent progress in generative methods, such as Marigold \cite{ke2024repurposing} and GeoWizard \cite{fu2025geowizard}, demonstrates that the use of diffusion priors can facilitate zero-shot depth estimation with improved prediction details and smoothness. However, despite their impressive performance in complex scenarios, these single-image depth estimation methods often face challenges in maintaining consistency across multiple frames.
% metric depth prediction, depth pro

\noindent \textbf{Video Depth.} Video depth estimation raises a challenge of consistency. To ensure frame-to-frame consistency in video depth prediction, previous studies have either directly predicted depth sequences on videos \cite{teed2018deepv2d,zhang2019exploiting,wang2023neural,yasarla2023mamo,yasarla2024futuredepth} or utilized online test-time optimization to tailor the network to overfit each video during inference \cite{chen2019self,kopf2021robust,luo2020consistent,zhang2021consistent}. Recent advancements in video diffusion models \cite{ho2022video,blattmann2023stable} have inspired their use as model priors for various video-related tasks, including video depth estimation. By treating the input video as the condition, ChronoDepth \cite{shao2024chrono}, DepthCrafter \cite{hu2024depthcrafter}, and Depth Any Video \cite{yang2024depth} leverage the video diffusion prior for high-fidelity, consistent video depth estimation. However, these methods still suffer from inconsistencies between sliding windows during long video inference and lack guarantees for geometric consistency.
% TODO depthcrafter's feature
% TODO However, limited to a batch, cross window consistency is only assured by cross attention. Our work enables the geometry consistency via diffusion guidance

\noindent \textbf{Diffusion Guidance}. Guided diffusion adjusts the sampling process of a pre-trained diffusion model using feedback from guidance functions, enabling flexible, training-free control over outputs. These functions, which include linear operators \cite{wang2022zero}, non-linear functions \cite{chung2022diffusion}, and off-the-shelf tools incorporating text, features, or segmentation \cite{graikos2022diffusion, bansal2023universal}, allow precise content manipulation on generation \cite{kawar2022denoising, epstein2023diffusion, yu2023freedom, luo2024readout}. In this work, we use guidance to inject scale and geometric constraints into the diffusion process, achieving scale- and geometry-consistent video depth estimation, especially for long videos.
% Classifier guidance \cite{dhariwal2021diffusion} was initially proposed to train a classifier on images at different noise scales, using the classifier's gradients to guide the sampling process. 
\section{Method}
\textbf{Overview.} As shown in Fig. \ref{fig:overview}, our method works by refining the intermediate results at each denoising step during inference for diffusion-based video depth prediction methods. Following common practice, the video is split into overlapping windows and depth is sequentially predicted in a window-by-window manner. During denoising, after predicting the noise at the current timestep with diffusion model, our method first enforces geometry consistency within each window through geometry guidance, and for the subsequent windows, scale guidance is applied to synchronize depth scale across different windows.

In this section, we first provide background on diffusion guidance and diffusion-based video depth prediction methodologies in Sec.\ref{sec:bg}. Diffusion guidance serves as a key mechanism in our method, enabling the application of regularization during the diffusion-based video depth prediction process to enhance both scale and geometry consistency. Subsequently, in Sec.\ref{sec:scale} and Sec.\ref{sec:geo}, we present the details of our scale guidance and geometry guidance.

\subsection{Preliminaries}\label{sec:bg}
\subsubsection{Diffusion Guided Sampling}\label{sec:guidance}
Universal guidance \cite{bansal2023universal} expands upon classifier guidance \cite{dhariwal2021diffusion}, enabling diffusion models to be controlled by a variety of guidance modalities through forward guidance and backward guidance. Forward guidance is applied after noise prediction in each sampling step.
Firstly, a \textit{predicted} clean sample $\hat{z_{0}}$ is computed. For example, for DDIM \cite{song2020denoising}:
\begin{equation}\label{eq:pred_x0}
   \hat{z_{0}}=\frac{z_{t}-(\sqrt{1-\alpha_{t}})\epsilon_{\theta}(z_{t}, t)}{\sqrt{\alpha_{t}}} 
\end{equation}
Subsequently, guided sampling is performed by updating the predicted noise $\epsilon_{\theta}(z_{t}, t)$ with the gradient of loss function $\mathcal{L}$:
\begin{equation} \label{eq:forward}
    \Hat{\epsilon}_{\theta}(z_{t}, t) = \epsilon_{\theta}(z_{t}, t) + s(t) \cdot \nabla_{z_{t}}\mathcal{L}(c, f(\hat{z_{0}}))
\end{equation}
where $s(t)$ regulates the guidance strength and $c$ denotes the condition signal for guidance. The loss function $\mathcal{L}$ measures the compatibility between the signal $c$ and the current predicted clean sample $\hat{z_{0}}$. A lower value indicates better compatibility. Applying guidance steers the generated samples in a direction that satisfies the constraints of $c$ by regularizing the noise prediction in the denoising process.

Backward guidance directly optimizes $\hat{z_{0}}$ with the constraints with gradient descent and obtains an optimized sample $\hat{z_{0}}+\Delta z_0$. Then the noise prediction is updated by substituting $\hat{z_{0}}+\Delta z_0$ into Eqn.(\ref{eq:pred_x0}). Backward guidance prioritizes the enforcement of constraints, while the forward guidance applies a smoother and more gradual form of guidance.

\subsubsection{Video Depth Diffusion}
Generative video depth estimation is framed as a conditional diffusion generation problem in a video-to-video fashion to model the distribution $p(\mathbf{d}|\mathbf{v})$, where $\mathbf{v} \in \mathbb{R}^{T\times H\times W\times 3}$ represents an input video and $\mathbf{d} \in \mathbb{R}^{T\times H\times W}$ represents the predicted depths \cite{hu2024depthcrafter,shao2024chrono,yang2024depth}. 
The models are trained based upon the image-to-video diffusion model Stable Video Diffusion (SVD) \cite{blattmann2023stable} with modifications to adapt to depth estimation. 
The conditional input to the denoiser is changed to the encoded latents of a whole video $\mathbf{z}^{(\mathbf{v})}$ rather than a single image as SVD. 
Meanwhile, to encode the depth sequences to the latent space, the depth map is normalized to \textit{affine-invariant} depth per sequence and replicated three times to encode to the latent space via a Variational Auto Encoder (VAE) \cite{kingma2013auto} to obtain depth latents $\mathbf{z}^{(\mathbf{d})}$. At inference, the denoised latent is decoded and then averaged across three channels to obtain the predicted depth.

\subsection{Scale Guidance for Cross-Window Consistency} \label{sec:scale}
When predicting video depth in a single forward pass, consistency between multiple frames is naturally ensured by cross-attention modules in the diffusion network. However, due to computational resource limitations, it is often impractical to process an entire video, which can consist of hundreds of frames, in a single pass to ensure cross-frame consistency. To address this challenge, previous methods \cite{shao2024chrono, hu2024depthcrafter} commonly split the video into overlapping sliding windows and predict depth sequentially for each window.

\begin{figure}[t!]
    \centering
    \includegraphics[width=0.5\textwidth]{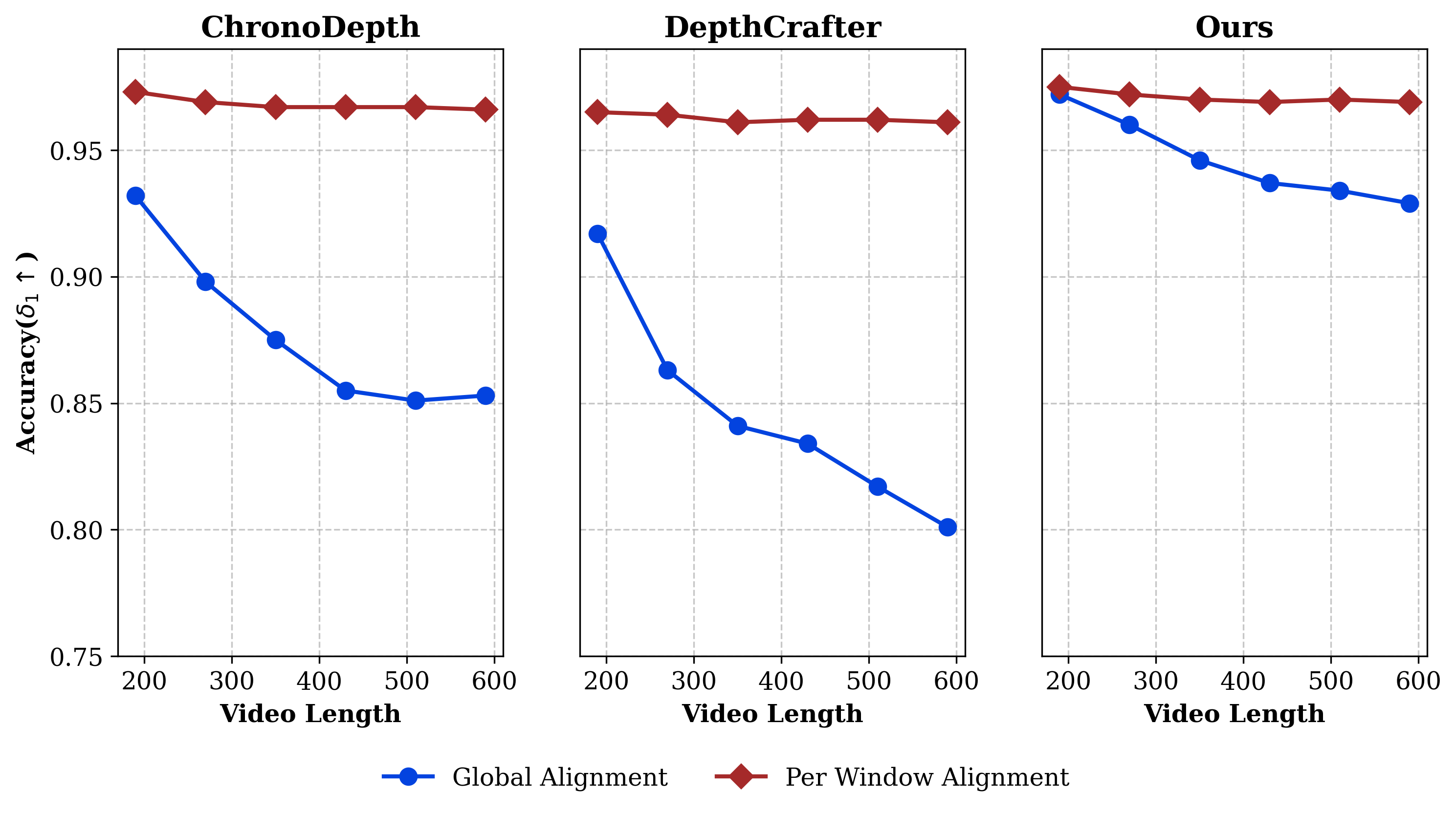}
    \caption{\textbf{Existing methods suffer from scale inconsistency across windows as video length increases.} We evaluate the same depth predictions on Bonn using two alignment strategies: aligning with a shared scale and shift factor for entire video versus per-window alignment. The shared alignment causes a significant accuracy drop, highlighting large scale variations between windows.}
    \label{fig:perwindow}
    \vspace{-0.5cm}
\end{figure}

To ensure consistency between different windows, DepthCrafter \cite{hu2024depthcrafter} employs diffusion inversion \cite{song2020denoising}, initializing the noise of the overlapping region by directly adding noise to the denoised latent from the previous window. The subsequent frames in the window are initialized from random Gaussian noise. Additionally, the overlapping region in the current denoised latent is further interpolated with the denoised latent from the previous window. Meanwhile, ChronoDepth \cite{shao2024chrono} proposes an approach where distinct noise levels are independently sampled for each frame within the window during both training and inference. The overlapping frames are directly initialized using the previous depth frames with a minimal level of noise.

In these existing solutions, depth scale information from previous windows is injected into the current window through the overlapping region at the beginning of the diffusion process and propagates to subsequent frames via cross-attention in the diffusion network. However, as shown by our experimental results in Table \ref{tab:long_depth}, these initialization-based strategies exhibit limited effectiveness in maintaining depth consistency across windows. Following the practice in affine-invariant video depth estimation \cite{hu2024depthcrafter}, before evaluation, we align the estimated depths with the ground truth using a shared scale and shift, optimized via least squares, across the entire video. Under this configuration, multiple sliding windows share a global scaling factor. In Table \ref{tab:long_depth}, we observe a significant decline in depth estimation accuracy for previous methods \cite{shao2024chrono, hu2024depthcrafter} as the video length and the number of sliding windows increase.

We further conduct a case study on the Bonn dataset \cite{palazzolo2019iros}, evaluating the same depth predictions using an alternative strategy that aligns depth individually for each window. This approach optimizes the scale and shift for each window independently, rather than applying a global scale and shift factor across all windows. As illustrated in Fig. \ref{fig:perwindow}, per-window alignment significantly improves depth estimation accuracy by a large margin, highlighting that depth scale variations between windows become more pronounced as the number of sliding windows increases. This leads to degraded depth estimation performance for long videos, indicating the scale information injected at diffusion initialization is lost during the denoising process.

% our proposed forward scale guidance method
To handle this cross-window consistency issue, we propose to exert depth scale synchronization with the previous window \textbf{during} the denoising process rather than just in the diffusion initialization. We implement such synchronization regularization via forward guidance. We store the denoised latent ${z_{0, \mathrm{last}}}$ from the previous window. During each denoising step, we predict the clean latent $\hat{z_{0, \mathrm{cur}}}$ using Eqn.~(\ref{eq:pred_x0}) and decode both the predicted latent $\hat{z_{0, \mathrm{cur}}}$ and the preceding latent ${z_{0, \mathrm{last}}}$ into depth maps $\hat{d_\mathrm{cur}}$ and $d_\mathrm{last}$ respectively. We then compute a scale $s$ and shift $t$ via least squares to align the depths of the overlapping frames from the two windows. Then the aligned depth over the entire window is computed as:
\begin{equation}
    d_\mathrm{aligned}=s\cdot \hat{d_\mathrm{cur}}+t
\end{equation}
We encode $d_\mathrm{aligned}$ to latent space to obtain aligned latent $z_{0, \mathrm{aligned}}$, which is used as a pseudo-label to constrain the predicted latent via an MSE loss:
\begin{equation}
    \mathcal{L}_{scale}=\frac{1}{n}\sum_{i=1}^{n}{(z_{0, \mathrm{aligned}, i}-\hat{z_{0, \mathrm{cur},i}})^2}
\end{equation}
The loss gradients are back-propagated to update the noise prediction according to Eqn.~(\ref{eq:forward}). To ensure computational efficiency, the guidance and backpropagation are performed directly in the latent space. As a result, we employ forward guidance for this latent-level regularization, as it provides smoother regularization compared to backward guidance.

Compared to existing initialization-based injection methods, our scale guidance strategy applies regularization to the current window throughout the entire diffusion process, rather than solely at the beginning of the diffusion denoising process. This approach achieves stronger regularization to synchronize depth scales across windows.

\subsection{Geometry Guidance for Intra-Window Consistency} \label{sec:geo}
Existing diffusion-based video depth estimation methods use 2D pre-trained priors and cross-attention modules in the diffusion network to ensure appearance consistency of predicted depth maps within a single window. However, unlike 2D tasks, depth describes 3D scene structures, and 2D priors inherently lack the geometric awareness required for accurate 3D geometry. As illustrated in Fig. \ref{fig:teaser}, the depth maps may appear correct in 2D visualizations, but the corresponding 3D structures often exhibit 3D inaccuracies. 

Moreover, monocular video frames, along with their corresponding depth maps, provide a comprehensive representation of the scene geometry. Each frame can generate a segment of the scene's point cloud when projected with its depth. Since these frames represent the same scene, their projections should align seamlessly, establishing an inherent geometric constraint between depth maps. However, existing consistency schemes only enable coarse alignment between frames, which is insufficient to accurately match the underlying geometry. For instance, as shown in Fig. Y, the projected walls are staggered and misaligned in 3D space, indicating a lack of geometric consistency.

To address the aforementioned issues, we propose a geometry guidance mechanism to ensure geometric correctness and consistency between frames within each sliding window using backward diffusion guidance. First, we compute the predicted clean latent $\hat{z_{0}}$ at the current noise level using Eqn. (\ref{eq:pred_x0}). The latent $\hat{z_{0}}$ is then decoded into image space to obtain the predicted depth. Next, we apply multiple geometry constraints to regularize and optimize this depth into a geometrically accurate version. The updated depth is subsequently encoded back into the latent space, and backward guidance is employed to update the noise prediction in the current denoising loop, as described in Sec. \ref{sec:guidance}.

Specifically, with the assistance of an off-the-shelf tracking prediction network \cite{karaev2024cotracker3}, we estimate the camera transformation from the RGB video and its corresponding depth frames using the Perspective-n-Point (PnP) method \cite{fischler1981random}. We then compute a depth reprojection loss and a tracking loss to ensure that different depth frames are aligned in 3D space, thereby enforcing geometric consistency between depth frames. The depth reprojection loss operates in a pairwise manner between two views, projecting one depth frame onto another and minimizing the difference between the projected depth and the depth at the target view. The tracking loss projects tracked points from different frames into 3D space using the predicted depth and minimizes their 3D distances. While the reprojection loss enforces projection consistency of the same 3D point across different views, the tracking loss additionally ensures consistency between pose estimation and depth across adjacent frames, forming a geometric optimization loop.

Additionally, with the assistance of an off-the-shelf, state-of-the-art diffusion-prior-based surface normal prediction model \cite{ye2024stablenormal}, we derive surface normals from the currently predicted depth and employ a surface normal loss to align with the results predicted by \cite{ye2024stablenormal}. Furthermore, we utilize an edge-aware smoothness loss $\mathcal{L}_{s}$ to promote local smoothness as proposed in \cite{heise2013pm}. %, to encourage local smoothness in the predicted depth.
% . This loss computes an L1 penalty on depth gradients, weighted by image gradients,

The overall guidance function for depth optimization is formulated as a weighted sum of these multiple loss terms::
\begin{equation}
\mathcal{L}=\alpha_{d}\mathcal{L}_{d}+\alpha_{t}\mathcal{L}_{t}+\alpha_{n}\mathcal{L}_{n}+\alpha_{s}\mathcal{L}_{s}
\end{equation}

We assign the largest weight to the depth reprojection term, as it plays the most critical role in ensuring geometric consistency. For detailed formulations and implementations of each loss term, please refer to the supplementary.

We employ backward guidance for geometric regularization, as the optimization process must operate on the decoded depth rather than the diffusion latent. Unlike forward guidance, which propagates gradients back to the current timestep's noise prediction in latent space, backward guidance directly optimizes the decoded depth in image space. This approach avoids the need for gradient backpropagation through the VAE decoding process, resulting in greater computational efficiency.

%------------------
% preliminary

% scale (important)
% existed issues, streaming, long video
% cross window scale not align

% scale alignment guidance implementation
% start step and end step detail

% discuss advantage compared to chronodepth and depthcrafter
% depthcrafter, latent inverse strategy
% chronodepth

% 
% optional: comparison with post optimization 

% geometry
% issue
% solution
% appearance consistency and geometry consistency
% method (shorten ver of cvpr submission)

% summary, cross window and inside window
\section{Experiments}
\begin{table*}[t!]
    \centering
    \scalebox{0.75}{
    \begin{tabular}{lccccccccccccc}
        \toprule
         \multicolumn{2}{c}{Frames per Window} & & \multicolumn{4}{c}{90} & & \multicolumn{4}{c}{110} & & 90 \\
         \cmidrule(lr){1-2} \cmidrule(lr){4-7} \cmidrule{9-12} \cmidrule{14-14}
         \multirow{2}{*}{Method} & \multirow{2}{*}{Type} & Video & \multicolumn{2}{c}{ScanNet} & \multicolumn{2}{c}{GMU KITCHEN} & Video & \multicolumn{2}{c}{KITTI} & \multicolumn{2}{c}{Bonn} & Video & ScanNet \\
        \cmidrule(lr){4-5} \cmidrule(lr){6-7} \cmidrule(lr){9-10}\cmidrule(lr){11-12} \cmidrule{14-14}
         & & Length & AbsRel$\downarrow$ & $\delta_{1} \uparrow$  & AbsRel$\downarrow$ & $\delta_{1} \uparrow$ & Length & AbsRel$\downarrow$ & $\delta_{1} \uparrow$ &  AbsRel$\downarrow$ & $\delta_{1} \uparrow$ & Length & MFC$\downarrow$\\
        \midrule
        DepthAnythingV2 & S & \multirow{6}{*}{150} & 0.148 & 0.776 & 0.188 & 0.645 &  \multirow{6}{*}{190} & 0.150 & 0.788 & 0.102 & 0.916 & \multirow{6}{*}{150} & 0.040 \\
        Marigold & S & & 0.179 & 0.740 & 0.214 & 0.584 & & 0.143 & 0.820 & 0.098 & 0.932  & & 0.131 \\ 
        NVDS & V & & 0.199 & 0.647 & 0.231 & 0.542 & & 0.212 & 0.658 & 0.163 & 0.771  & & 0.047 \\
        ChronoDepth & V &  & 0.172 & 0.749 & 0.196 & 0.650 & &  0.178 & 0.733 & 0.092 & 0.932 & & 0.028 \\
        DepthCrafter & V & & 0.141 & 0.799 & 0.143 & 0.795 &  & 0.114 & 0.879 & 0.095 & 0.917 & & 0.024 \\
        \textbf{DepthSync(Ours)} & V & & \textbf{0.113} & \textbf{0.870} & \textbf{0.113} & \textbf{0.881} & & \textbf{0.110} & \textbf{0.887} & \textbf{0.069} & \textbf{0.972}  & & \textbf{0.019} \\
        \midrule
        DepthAnythingV2& S  & \multirow{6}{*}{210} & 0.160 & 0.743 & 0.189 & 0.652 &  \multirow{6}{*}{270} & 0.156 & 0.777 & 0.117 & 0.887  & \multirow{6}{*}{210} & 0.040 \\
        Marigold & S & & 0.191 & 0.710 & 0.221 & 0.598 & & 0.146 & 0.809 & 0.113 & 0.901  & & 0.143 \\
        NVDS & V& & 0.208 & 0.622 & 0.223 & 0.578 & & 0.217 & 0.647 & 0.176 & 0.741  & & 0.048 \\
        ChronoDepth& V & & 0.182 & 0.726 & 0.201 & 0.659 & & 0.185 & 0.707 & 0.112 & 0.898   & & 0.028 \\
        DepthCrafter & V& & 0.156 & 0.757 & 0.155 & 0.761 & & 0.116 &  0.867 & 0.121 & 0.863  & & 0.025 \\
        \textbf{DepthSync(Ours)} & V & & \textbf{0.127} & \textbf{0.836} & \textbf{0.113} & \textbf{0.884}  & & \textbf{0.111} & \textbf{0.882} & \textbf{0.077} & \textbf{0.960}  & & \textbf{0.021} \\
        \midrule
        DepthAnythingV2 & S & \multirow{6}{*}{270} & 0.161 & 0.738 & 0.181 & 0.681 &  \multirow{6}{*}{350} & 0.161 & 0.760 & 0.125 & 0.870  & \multirow{6}{*}{270} & 0.041 \\
        Marigold & S & & 0.195 & 0.699 & 0.219 & 0.613 & & 0.151 & 0.796 & 0.123 & 0.884  & & 0.148 \\
        NVDS & V& & 0.211 & 0.614 & 0.222 & 0.586 & & 0.219 & 0.643 & 0.180 & 0.722  & & 0.048 \\
        ChronoDepth& V &  & 0.184 & 0.716 & 0.201 & 0.659 & & 0.185 & 0.705 & 0.124 & 0.875  & & 0.029 \\
        DepthCrafter& V & & 0.161 & 0.745 & 0.154 & 0.764 & & 0.119 & 0.859 &  0.130 & 0.841  & & 0.025 \\
        \textbf{DepthSync(Ours)}& V & & \textbf{0.136} & \textbf{0.814} &\textbf{0.120} & \textbf{0.860} & & \textbf{0.112}  & \textbf{0.874} & \textbf{0.083} & \textbf{0.946}  & & \textbf{0.021}\\
        \midrule
        DepthAnythingV2& S &\multirow{6}{*}{330} & 0.164 & 0.729 & 0.191 & 0.655 &  \multirow{6}{*}{430} & 0.160 & 0.762 & 0.127 & 0.861  &\multirow{6}{*}{330} & 0.041 \\
        Marigold & S & & 0.198 & 0.691 & 0.228 & 0.611 & & 0.150 & 0.800 & 0.126 & 0.871  & & 0.166\\
        NVDS& V & & 0.214 & 0.608 & 0.229 & 0.568 & & 0.221 & 0.641 & 0.182 & 0.712  & & 0.049 \\
        ChronoDepth& V & & 0.187 & 0.709 & 0.217 & 0.621 & & 0.186 & 0.706 & 0.130 & 0.855  & & 0.029 \\
        DepthCrafter & V & & 0.168 & 0.725 & 0.153 & 0.756 & & 0.123 & 0.853 & 0.134 & 0.834  & & 0.025 \\
        \textbf{DepthSync(Ours)}& V & & \textbf{0.138} & \textbf{0.802} & \textbf{0.128} & \textbf{0.842} &  & \textbf{0.115} & \textbf{0.871} & \textbf{0.090} & \textbf{0.937}  & & \textbf{0.021}\\
        \midrule
        DepthAnythingV2 & S & \multirow{6}{*}{390} & 0.167 & 0.724 & 0.193 & 0.645 &  \multirow{6}{*}{510} & 0.158 & 0.766 & 0.128 & 0.861  & \multirow{6}{*}{390} & 0.041 \\
        Marigold & S & & 0.202 & 0.689 & 0.228 & 0.613 &  & 0.149 & 0.800 & 0.126 & 0.871 & & 0.148 \\
        NVDS & V& & 0.216 & 0.603 & 0.232 & 0.561 & & 0.221 & 0.641 & 0.185 & 0.701  & & 0.050 \\
        ChronoDepth & V & &  0.190 & 0.704 & 0.222 & 0.619 & & 0.186 & 0.705 & 0.132 & 0.851   & & 0.030\\
        DepthCrafter & V & & 0.169 & 0.721 & 0.161 & 0.746 & & 0.124 & 0.850 & 0.138 & 0.817  & & 0.025 \\
        \textbf{DepthSync(Ours)} & V & & \textbf{0.150} & \textbf{0.780} & \textbf{0.128} & \textbf{0.848} & & \textbf{0.116} & \textbf{0.870} & \textbf{0.091} & \textbf{0.934}  & & \textbf{0.021}\\
        \midrule
        DepthAnythingV2 & S & \multirow{6}{*}{450} & 0.169 & 0.722 & 0.193 & 0.644 &  \multirow{6}{*}{590} & 0.158 & 0.769 & 0.131 & 0.856  & \multirow{6}{*}{450} & 0.041 \\
        Marigold & S & & 0.205 & 0.686 & 0.241 & 0.597 & & 0.148 & 0.803 & 0.128 & 0.865  & & 0.147 \\
        NVDS & V & & 0.219 & 0.597 & 0.235 & 0.558 & & 0.223 & 0.640 & 0.189 & 0.692  & & 0.050   \\
        ChronoDepth & V & & 0.193 & 0.699 & 0.225 & 0.612 & & 0.193 & 0.691 & 0.132 & 0.853  & & 0.030 \\
        DepthCrafter & V & & 0.171 & 0.716 & 0.160 & 0.748 & & 0.173 & 0.727 & 0.143 & 0.801  & & 0.025\\
        \textbf{DepthSync(Ours)} & V & & \textbf{0.154} & \textbf{0.769} & \textbf{0.131} & \textbf{0.837} & & \textbf{0.117} & \textbf{0.869} & \textbf{0.093} & \textbf{0.929}  & & \textbf{0.021} \\
        % \midrule
        % \multirow{4}{*}{510} & NVDS & & & & & & &  \multirow{4}{*}{190} & \\
        % & ChronoDepth & \\
        % & DepthCrafter & \\
        % & \textbf{DepthSync(Ours)} & \\
        
        \bottomrule    
        
    \end{tabular}}
    \caption{\textbf{Depth Evaluation Results.} We compare our method with representative single-image (Type S) and video depth estimation models (Type V). Our approach significantly improves long-term video depth estimation, demonstrating enhanced consistency.}
    \vspace{-0.5cm}
\label{tab:long_depth}
\end{table*}

\begin{table}[t!]
    \centering
    \scalebox{0.7}{
    \begin{tabular}{ccccccc}
        \toprule
         Length & Metric & RelPose++ & PoseDiff & RayDiff & DC & \textbf{Ours}  \\
         \midrule
         \multirow{3}{*}{150} & ATE\ (m)$\downarrow$ & 0.604 & 0.570 & 0.522 & 0.154 & \textbf{0.144} \\
         & RPE trans\ (m)$\downarrow$ & 0.078 & 0.132 & 0.176 & 0.021 & \textbf{0.020} \\
         & RPE rot \ (deg)$\downarrow$ & 2.91 & 3.41 & 26.9 & 0.622 & \textbf{0.611} \\
         \midrule
         \multirow{3}{*}{210} & ATE\ (m)$\downarrow$ & 0.741 & 0.737 & 0.666 & 0.211 & \textbf{0.197} \\
         & RPE trans\ (m)$\downarrow$ & 0.093 & 0.149 & 0.197 &  0.022 & \textbf{0.020} \\
         & RPE rot \ (deg)$\downarrow$ & 2.96 & 4.09 & 28.3 & 0.624 & \textbf{0.620} \\
         \midrule
         \multirow{3}{*}{270} &ATE\ (m)$\downarrow$ & 0.846 & 0.854 & 0.792 & 0.256 & \textbf{0.242} \\
         & RPE trans\ (m)$\downarrow$ & 0.098 & 0.156   & 0.198 & 0.023 & \textbf{0.021} \\
         & RPE rot \ (deg)$\downarrow$ & 2.96 & 4.11 & 29.4 & 0.641 & \textbf{0.635} \\
         \midrule
         \multirow{3}{*}{330} & ATE\ (m)$\downarrow$  & 0.899 & 0.906 & 0.849 & 0.292 & \textbf{0.275} \\
         & RPE trans\ (m)$\downarrow$ & 0.093 & 0.158 & 0.200 & 0.023 & \textbf{0.021} \\
         & RPE rot \ (deg)$\downarrow$ & 2.97 & 4.11 & 30.3 & 0.642 & \textbf{0.627}\\
         \midrule
         \multirow{3}{*}{390} & ATE\ (m)$\downarrow$ & 0.941 & 0.954 & 0.891 & 0.334 & \textbf{0.316} \\
         & RPE trans\ (m)$\downarrow$ & 0.092 & 0.152& 0.204 & 0.023 & \textbf{0.021} \\
         & RPE rot \ (deg)$\downarrow$ & 2.99 & 4.12 & 30.5 & 0.650 & \textbf{0.643} \\
         \midrule
         \multirow{3}{*}{450} & ATE\ (m)$\downarrow$ & 1.00 & 1.01 & 0.942 & 0.367 & \textbf{0.350} \\
         & RPE trans\ (m)$\downarrow$ & 0.094 & 0.162  & 0.209 & 0.023 & \textbf{0.021} \\
         & RPE rot \ (deg)$\downarrow$ & 3.12 & 4.29 & 30.7 & 0.676 & \textbf{0.669} \\
         \bottomrule
    \end{tabular}
    }
    \caption{\textbf{Evaluation of Derived Poses from Our Predicted Depths.} Poses derived from our depths outperforms representative deterministic and generative pose estimation methods on ScanNet.}
\vspace{-0.2cm}
    
\label{tab:long_pose}
\end{table}

% \begin{table}[t!]
% \scalebox{0.8}{
%     \centering
%     \begin{tabular}{l|cc|cc}
%         \toprule
%         {\multirow{2}{*}{Method}} &
%         \multicolumn{2}{c|}{Depth} & \multicolumn{2}{c}{Pose} \\
%         \cmidrule{2-5} 
%         {}  & AbsRel $\downarrow$ & {$\delta<1.25$ $\uparrow$} & RPE (t)\ (m)$\downarrow$ & RPE (r) \ (deg)$\downarrow$ \\
%         \midrule
%         Post-Opt. & 0.108 & 0.885 & 0.090 &  3.510 \\ 
%         \textbf{Ours}  & \textbf{0.106} & \textbf{0.891} & \textbf{0.019} & \textbf{0.588} \\
        
%         \bottomrule    
%     \end{tabular}}
%     \caption{\textbf{Comparison with Post-Optimization on ScanNet.} Conducting geometric optimization in each diffusion sampling steps  yields better 3D structures than executing optimization after all diffusion steps, indicating an effective synergy between diffusion priors and geometric optimization in our proposed method.}
% \label{tab:abl_opt}
% \end{table}

\subsection{Implementation}
\textbf{Guidance Implementation.} We implement our DepthSync based on the state-of-the-art diffusion-based video depth prediction framework, DepthCrafter \cite{hu2024depthcrafter}. Aligned with DepthCrafter, we use the Euler scheduler \cite{karras2022elucidating} with 5 denoising steps, applying our guidance in the last 2 steps to balance efficiency and effectiveness. All experiments are conducted on a single NVIDIA A100-SXM4-40GB GPU. For more details please refer to the supplementary.

\noindent\textbf{Evaluation Protocol.} We adopt a more challenging protocol as per DepthCrafter \cite{hu2024depthcrafter}, optimizing a shared scale and shift across the entire video using least squares. This global alignment better reflects the depth consistency between video frames.
We evaluate depth accuracy using Absolute Relative Error (AbsRel$\downarrow$, $|\hat{\mathrm{d}}-\mathrm{d}|/\mathrm{d}$) and the prediction threshold accuracy ($\delta_1\uparrow$, percentage of $\mathrm{max}(\hat{\mathrm{d}}/\mathrm{d}, \mathrm{d}/\hat{\mathrm{d}})<1.25$). To measure the consistency, we adopt the multi-frame consistency (MFC) metric as per previous works \cite{shao2024chrono, luo2020consistent}. MFC measures the consistency between frames by warping the depth prediction from one frame to its adjacent with ground truth camera pose transformation, and evaluating the difference between the warped and target depths.

\noindent\textbf{Dataset}. We evaluate on four benchmarks: ScanNet v2 \cite{dai2017scannet} (100 scenes), GMU Kitchen \cite{georgakis2016multiview} (9 scenes), Bonn RGB-D Dynamic \cite{palazzolo2019iros} (26 scenes), and KITTI \cite{geiger2012we} (13 scenes), covering static to dynamic and indoor to outdoor scenes. Following \cite{hu2024depthcrafter}, we use  windows of 90 frames for ScanNet and GMU Kitchen, and 110 frames for Bonn and KITTI.

% TODO pose evaluation protocol?

\subsection{Zero-Shot Depth Estimation Results}\label{sec:eval}
We compare our method with several representative depth estimation approaches, including the single-frame deterministic method Depth Anything V2 \cite{yang2024depthanything}, single-frame generative methods Marigold \cite{ke2024repurposing}, and video-based depth estimation methods, including NVDS \cite{NVDS}, ChronoDepth \cite{shao2024chrono}, and DepthCrafter \cite{hu2024depthcrafter} in Table \ref{tab:long_depth}.

\noindent\textbf{Scale Consistency.}
To evaluate scale consistency, we test across six video length settings, corresponding to 2 to 7 overlapping windows during inference on each dataset. As shown in Table \ref{tab:long_depth}, our method consistently outperforms baselines across all lengths and datasets. Baseline methods exhibit significant accuracy declines as video length increases, primarily due to scale inconsistency between windows (Fig. \ref{fig:perwindow}). In contrast, our scale guidance strategy synchronizes scales between adjacent windows, effectively mitigating this issue and enhancing consistency. Since the Bonn dataset contains a large number of highly dynamic objects and humans in the scene, only scale guidance is applied to this dataset. Thus the performance gain observed on the Bonn dataset is solely attributable to the scale guidance. Notably, compared to DepthCrafter, to which our guidance is applied, our approach achieves a significant improvement of approximately 16\% in the depth accuracy metric $\delta_1$ for videos with lengths of up to 590 frames on Bonn.
 
\noindent\textbf{Geometry Consistency.}  
Since the computation of the MFC (Multi-Frame Consistency) metric requires ground truth poses for each frame, we evaluate it on ScanNet. Our method consistently achieves lower MFC values across all video lengths, demonstrating superior geometry consistency compared to baseline approaches. As MFC is computed between adjacent frames and averaged over the video, its value varies only slightly with increasing video length.

Given that video depths inherently represent 3D structure, we further evaluate the geometric quality by assessing poses derived from our depths. Using pixel correspondences between adjacent frames predicted from an offline tracking predictor \cite{karaev2024cotracker3}, we compute poses via PnP-RANSAC and evaluate them using visual odometry metrics: absolute translation error (ATE), relative translation error (RPE trans), and relative rotation error (RPE rot) \cite{zhang2022structure}. As shown in Table~\ref{tab:long_pose}, the poses derived from our depths outperform those from the deterministic method RelPose++ \cite{lin2024relpose++}, the generative methods PoseDiffusion \cite{wang2023posediffusion} and RayDiffusion \cite{zhang2024cameras} as well as the poses derived from the baseline method DepthCrafter. The results for RayDiffusion are reported from the model trained on CO3Dv2, as it is the only open-sourced pretrained model available. These results validate the geometric accuracy of our predicted depths.

\begin{figure*}[h]
    \centering
    \includegraphics[width=0.98\textwidth]{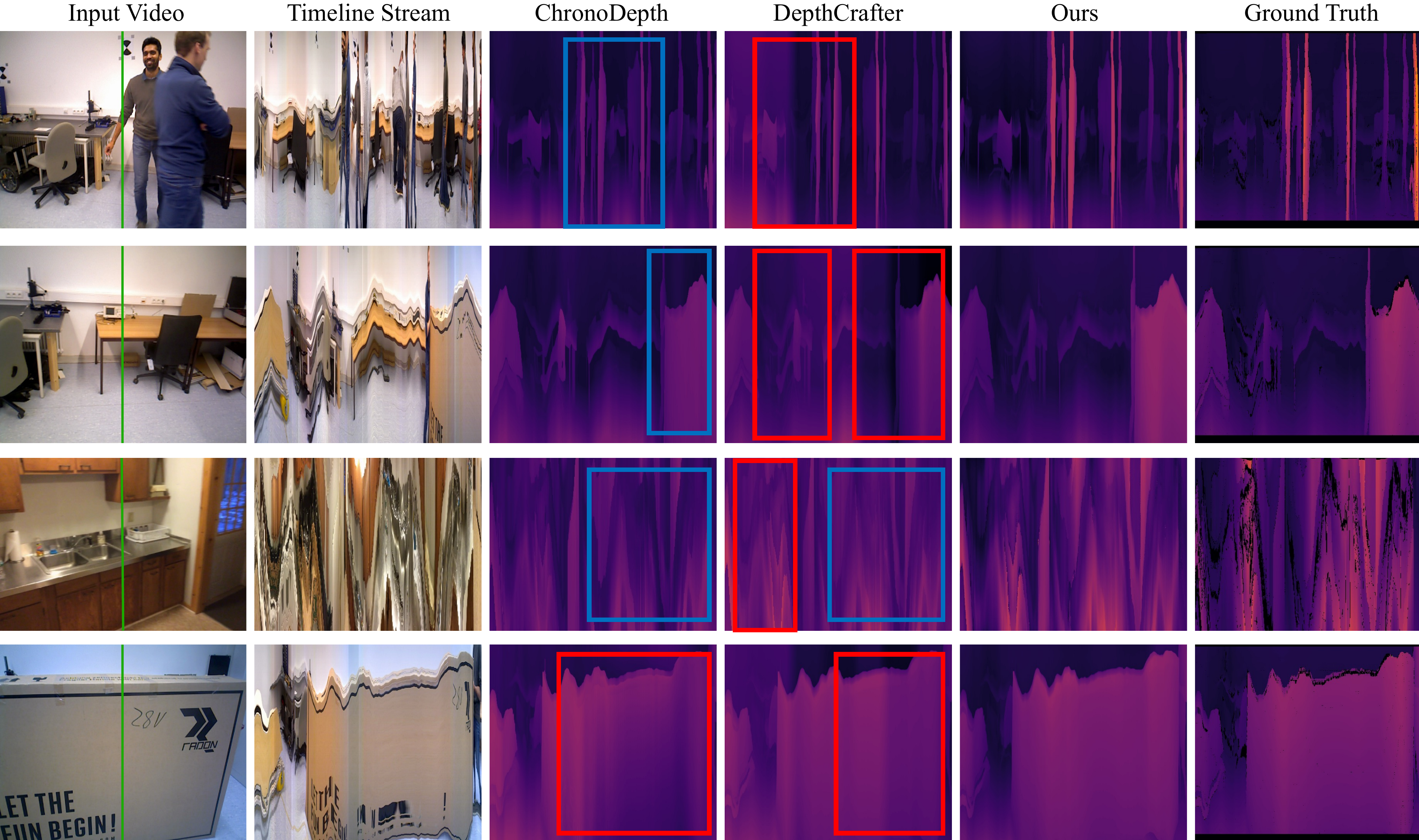}
    \caption{\textbf{Qualitative Comparison for Long Video Depth Estimation.} We compare predictions on 450-frame ScanNet videos and 590-frame Bonn videos. Slicing along the timeline at the green-line position (first column), we concatenate results to visualize temporal changes in color and depth. Red boxes highlight depth scale inconsistencies, while blue boxes mark depth inaccuracies in previous methods. Our method shows superior depth accuracy and scale consistency over time. Depth error map comparisons are provided in Appendix.}
    
    \label{fig:timeseq}
    \vspace{-0.35cm}
\end{figure*}

\begin{figure*}[h]
    \centering
    \includegraphics[width=\textwidth]{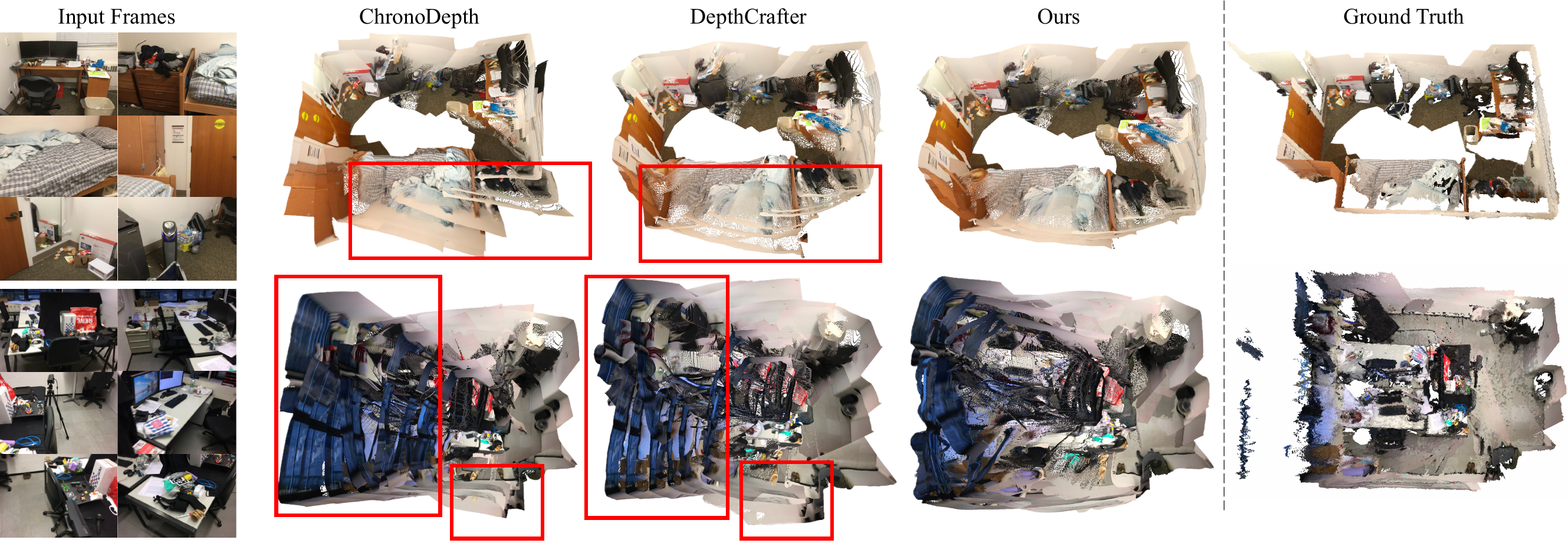}
    \vspace{-0.5cm}
    \caption{\textbf{Qualitative Reconstruction Comparison on ScanNet.} We project video depth into 3D space, demonstrating that our method produces more accurate geometry. Red boxes highlight misaligned walls in previous methods' reconstructions.
    }
    \label{fig:reconstruction}
    \vspace{-0.3cm}
\end{figure*}

\noindent\textbf{Qualitative Results.}
We compare our method with two representative diffusion-based video depth estimation methods: ChronoDepth \cite{shao2024chrono} and DepthCrafter \cite{hu2024depthcrafter}. Fig. \ref{fig:timeseq} visualizes the results for long videos. The second column displays the temporal stream of images, obtained by slicing along the timeline at the green-line position marked in the first column. Before visualization, we align the disparity with the scale of ground truth. The red boxes highlight inconsistencies in depth scale over time, while the blue boxes indicate inaccuracies in depth values. Our method demonstrates superior depth accuracy and greater scale consistency over time compared to previous methods. Additionally, we project the predicted video depth on ScanNet into 3D space to generate reconstruction results with ground truth poses. As shown in Fig. \ref{fig:reconstruction}, ChronoDepth and DepthCrafter produce walls that are staggered in 3D space, whereas our method reconstructs walls from different frames with significantly better alignment, validating the improved geometric consistency of our approach.

% The quantitative results demonstrate that while DepthCrafter achieves remarkable temporal consistency as a state-of-the-art video depth generative model, it struggles to fully perceive the 3D structure of scenes and ensure geometric consistency across frames. We also provide qualitative reconstruction visualizations comparisons. As shown in Fig.~\ref{fig:qua_cp}, our DepthSync shows better capability to recover accurate depth aligned with the 3D geometric structure compared to the purely diffusion-based DepthCrafter. For example, in the left bottom example in Fig.~\ref{fig:qua_cp}, the wall reconstructed by DepthCrafter's depth are curved when observing from the side view, while the wall reconstructed by our method is more plain. In the right bottom cases, the cabinet reconstructed by DepthCrafter is severely deformed while our method is able to recover the rigid geometric structure. On the overhand, the as shown in the two top cases in Fig.~\ref{fig:qua_cp}, the walls reconstructed from different frames are more closer by our method than DepthCrafter, showcasing that besides geometric accuracy, DepthSync is also able to enhance geometric consistency across frames. More results are in supplementary materials.

\subsection{Ablation Study}
In this section, we present the experimental results of an ablation study on each guidance term, along with a comparison to post-optimization methods, as detailed in Table \ref{tab:abl}. Additionally, we provide ablation results in the supplementary materials to evaluate the effectiveness of each loss term in the geometry guidance and the optimal step to apply the guidance process. To ensure a focused and effective analysis, all ablation experiments are conducted on the first 35 scenes of the ScanNet test set. We maintain a window size of 90 frames, with an overlap of 30 frames between consecutive windows, and evaluate the first 150 frames of each video. Under this configuration, the depth for each video is inferred from 2 overlapping sliding windows.

\begin{table}[t!]
    \centering
    \scalebox{0.9}{
    \begin{tabular}{lccc}
        \toprule
        Strategy &  AbsRel$\downarrow$ & $\delta_{1} \uparrow$ & MFC  \\
        \midrule
        \textbf{Baseline} & 0.174 & 0.702 & 0.024 \\
        \midrule 
        \multicolumn{4}{l}{\textbf{Guidance Term Ablation}} \\
        Scale Guidance Only & 0.151 & 0.772 & 0.024 \\
        Geometry Guidance Only & 0.166 & 0.720 & 0.018\\
        \midrule
        \multicolumn{4}{l}{\textbf{Post Optimization}} \\
        Post Scale Alignment & 0.157 & 0.750 & 0.024\\
        Post Geometry Optimization & 0.175 & 0.694 & 0.020 \\
        Post Scale \& Geometry & 0.151 & 0.757 & 0.020 \\
        \midrule
        \textbf{Ours} & \textbf{0.137} & \textbf{0.801} & \textbf{0.018} \\
        % \midrule
        % Baseline 210 & 0.191 & 0.654 \\
        % post align scale 210 & 0.175 & 0.702 \\
        % scale guidance only 210 & 1.170 & 0.725 \\
        % geo only 210 & 0.208 & 0.623 \\
        % long 210 v26 & 0.154 & 0.761 \\
        \bottomrule    
        
    \end{tabular} }
        
   \caption{\textbf{Ablation Studies on Each Guidance Term and Comparison with Post Optimization.} The two terms work synergistically, achieving better performance when combined than individual use. Our guidance strategy, which applies constraints during the denoising process, outperforms post-denoising regularization.}
\label{tab:abl}
\vspace{-0.3cm}
\end{table}

\noindent \textbf{Effectiveness of Each Guidance Term.}
We first conduct an ablation study to validate the effectiveness of each component of our method. As shown in Table~\ref{tab:abl}, applying scale guidance alone results in a remarkable 13.2\% reduction in relative depth errors and an 8.73\% improvement in depth accuracy, highlighting that scale variance between windows is a significant issue in existing methods. In contrast, applying geometry guidance alone yields more detailed improvements, reducing the MFC value from 0.024 to 0.018. Combining both guidance terms achieves the best results, with a 21.3\% reduction in relative errors and a 14.1\% improvement in accuracy, demonstrating that the two terms complement each other and work synergistically to enhance performance.

\noindent \textbf{Comparison with Purely Post-Optimization.}
We further compare our diffusion guidance strategy with post-optimization methods, which apply optimization after the model's inference process. The post-optimization strategies apply the same regularization terms as our method, but the key difference lies in where these regularizations are injected: either entirely after the diffusion denoising steps or integrated within the denoising loop. As shown in Table~\ref{tab:abl}, while post-optimization strategies improve the baseline method, our guidance strategy achieves superior results. Meanwhile, applying each guidance term individually consistently yields better results compared to applying the same terms in a post-optimization framework. This demonstrates that integrating regularization terms within the denoising loop facilitates beneficial interactions between the regularization and the diffusion process, leading to better outcomes. By incorporating regularization during the denoising loop, the diffusion prior aids the optimization process in avoiding local minima and converging toward a more optimal solution.

\section{Conclusion}
In this paper, we propose DepthSync, a training-free, diffusion-based video depth estimation method designed to enhance both scale and geometry consistency for depth estimation, particularly for long videos. Our approach employs forward and backward diffusion guidance.
These guidance mechanisms enforce regularization to synchronize depth scale cross windows and ensure geometry alignment within each window throughout the diffusion denoising process. Extensive experimental results across various datasets demonstrate the effectiveness of our method in producing depth estimates with significantly improved scale and geometry consistency, especially for long videos.

\section*{Acknowledgments}
This work was supported by the National Natural Science Foundation of China (No. 62495061, 62361146854), and Tsinghua-Tencent Joint Laboratory for Internet Innovation Technology.

{
    \small
    \bibliographystyle{ieeenat_fullname}
    \bibliography{main}
}

\appendix

\maketitlesupplementary
\section{Implementation Details}
\subsection{Hyperparameters}
We employ the Euler scheduler \cite{karras2022elucidating} with 5 denoising steps, as per DepthCrafter \cite{hu2024depthcrafter}, we apply both guidance terms sequentially in the last two steps: geometry guidance first to refine geometry within each window, followed by scale guidance to synchronize scales across windows.

Scale guidance uses forward guidance \cite{bansal2023universal}, back-propagating the MSE loss gradients (Eqn.~(4) in the main paper) to the noise prediction. This process iterates 1000 times in the final denoising step. The base learning rate $\mathrm{lr}$ (Eqn. (2) in the main paper) is set to 6e4 for the second-to-last step and 1e6 for the final step, decaying as:
\begin{equation}
    s(t) = \mathrm{lr} \times 0.98^{\mathrm{iter}//100}
\end{equation}
Early termination occurs if the loss falls below $5e-4$.

Geometry guidance uses backward guidance, optimizing the predicted depth decoded from the clean latent (Eqn. (1) in the main paper). We set the base learning rate to 0.01 for the second-to-last step and 0.02 for the final step, with 80 optimization iterations per step. We use the AdamW optimizer (betas = (0.9, 0.999), weight decay = 1e-4) and a cosine annealing scheduler (eta min = base lr * 0.01, T max = 80).

\subsection{Loss Terms in Geometry Guidance}
We employ geometric optimization based on multiple geometric constraints as the geometry guidance, specifically, comprising of  global \textit{reconstruction constraints} and local \textit{detail constraints} to ensure geometric consistency. In each denoising step, we first obtain the current prediction of the clean latent using Eqn. (1) in the main paper. We then decode it back to the image space to obtain the predicted depth maps $\mathbf{d}_{i}$ on input video frames $\mathbf{I}_{i}$, which serve as inputs for the guidance loss computation. 

\subsubsection{Global Geometric Constraints constraints} 
A monocular video, along with its corresponding depth maps per frame, provides a comprehensive representation of the scene geometry. Specifically, each frame can generate a segment of the scene's point cloud  when projected with its depth. Different frames should form aligned projections, establishing an inherent constraint between depths.
Aligning these projections requires access to the camera pose of each frame. To determine the camera poses, we employ an off-the-shelf tracking prediction network \cite{karaev2024cotracker3} to generate pairwise dense tracking pixel correspondences between adjacent frames and estimate the camera poses $\mathbf{P}_{i}$ of each frame using Perspective-$\mathrm{n}$-Point (PnP)  \cite{fischler1981random}. We employ the  solvePnPRansac inferface from opencv-python library for PnP computation. With the determined poses, we apply depth reprojection loss and tracking loss to guarantee global consistency, ensuring that the 3D structure formed with the video depths within sliding window represents a geometrically consistent scene.

\noindent \textbf{Depth Reprojection Loss.} Given a frame $\mathbf{I}_{i}$, camera intrinsics $\mathbf{K}$ and  relative pose transformation $\mathbf{T}_{i\to j}$ to another frame $\mathbf{I}_{j}$, we project pixels in $\mathbf{I}_{i}$, denoted as $p_i$, to 3D points, transform them to the coordinates of $j$ and project to $\mathbf{I}_{j}$ to get the corresponding pixels $p_j$, with a pixel correspondence between two frames as:
\begin{equation}
    p_{j}\sim \mathbf{K}\mathbf{T}_{i\to j}\mathbf{d}_{i}(p_{i})\mathbf{K}^{-1}p_{i},
\end{equation}
Consequently, depth map $\mathbf{d}_{j}$ can be warped $\mathbf{d}_{j\to i}$ under this pixel correspondence relationship, and its difference to $\mathbf{d}_{i}$ forms the depth reprojection loss $\mathcal{L}_r$, which is computed as an L1 difference between the two depth maps:
\begin{equation}
    \mathcal{L}_d^{i, j} = |\mathbf{d}_{i}-\mathbf{d}_{j\to i}|
\end{equation}

The depth reprojection loss is computed and averaged across all pairs within frame distance of 3 in a sliding window. 
As for the case where camera intrinsic matrix $K$ is unknown, we can approximate it using image height (h) and width (w), with focal length as $(w+h)/2$ and principal point as $(w/2, h/2)$ as common practice.

\noindent \textbf{Tracking Loss.} Tracking correspondences between $\mathbf{I}_{i}$ and frame $\mathbf{I}_{j}$ allow us to project pixels to 3D using corresponding depth predictions, obtaining points $P_i$ and $P_j$. Transforming $P_j$ to coordinate $i$ using derived camera poses from PnP, the L2 difference to $P_i$ forms a tracking loss:
\begin{equation}
    \mathcal{L}_t^{i, j}=||P_{i}-P_{j\to i}||_{2}
\end{equation}
The tracking loss is computed for adjacent pairs only for simplicity.

\subsubsection{Local Geometric Constraints}
\noindent\textbf{Surface Normal Loss.} To improve the local structure of the generated depth, we employ an offline surface normal generation network \cite{ye2024stablenormal} to pre-compute surface normals $\mathbf{n}_{i}^{pre}$ for each video frame. We compute the surface normal $\mathbf{n}_{i}$ from the predicted depth map and align it to the direction of the pre-computation result with surface normal angular loss:
\begin{equation}
    \mathcal{L}_n^{i} = 1-{\mathrm{cos}(\mathbf{n}_{i}^{pre}}, \mathbf{n}_{i})
\end{equation}

\noindent \textbf{Smoothness Loss.} We also use edge-aware smoothness loss $\mathcal{L}_{s}$ to encourage local smoothness by compute an L1 penalty on depth gradients with image gradients as weight \cite{heise2013pm}:
\begin{equation}
    \mathcal{L}_{s}^{i} = |\partial_{x}\mathbf{d}_{i}|e^{-|\partial_{x}\mathbf{I}_{i}|} + |\partial_{y}\mathbf{d}_{i}|e^{-|\partial_{y}\mathbf{I}_{i}|}
\end{equation}
The detail constraints are computed from each frame respectively and averaged over the batch.

\subsubsection{Overall Loss}
The overall guidance function comprises a weighted sum of the multiple loss terms above:
$$
\mathcal{L}=\alpha_{d}\mathcal{L}_{d}+\alpha_{t}\mathcal{L}_{t}+\alpha_{n}\mathcal{L}_{n}+\alpha_{s}\mathcal{L}_{s}
$$
We assign the largest loss weight to the depth reprojection loss term as it plays the major role in aligning the geometry between frames. We set $\alpha_{d}=35, \alpha_{n}=0.1,\alpha_{s}=1, \alpha_{t}=2$ in all the experiments in the paper.
\section{Comparison with Reconstruction Methods}
We compare our method with representative reconstruction methods on the first 10 ScanNet test scenes. COLMAP\cite{schoenberger2016sfm} and 2DGS \cite{Huang2DGS2024} (uses COLMAP poses) fail on 3 scenes due to failed COLMAP pose estimation, while our method reconstructs all scenes successfully. For shared scenes, our approach achieves more accurate depth and poses (Table \ref{tab:recons}).

\begin{table}[ht]
    \centering
    \scalebox{0.9}{
    \begin{tabular}{l|ccc}
        \toprule
        Method & COLMAP \cite{schoenberger2016sfm} & 2DGS\cite{huang20242d} & \textbf{DepthSync} \\
        \midrule
        \# Failure Cases & 3 & 3 & \textbf{0} \\
        \midrule
        AbsRel$\downarrow$ & 0.464 & 0.136 & \textbf{0.098} \\
        $\delta<1.25$ $\uparrow$ & 0.479 & 0.823 & \textbf{0.910} \\
        \midrule
        ATE$\downarrow$ & \multicolumn{2}{c}{0.214} & \textbf{0.088} \\
        RPE t$\downarrow$ & \multicolumn{2}{c}{0.0914} & \textbf{0.0198}\\
        RPE r$\downarrow$ & \multicolumn{2}{c}{6.65} & \textbf{0.611}\\
        \bottomrule
        \end{tabular}}
    \caption{\textbf{Comparison with Reconstruction Methods.} Our method demonstrates better geometric accuracy as well as better robustness compared to conventional reconstruction methods.}
        
\label{tab:recons}
\end{table}

\section{Inference Cost and Supplementary Evaluation}
We report time and peak GPU memory usage during inference in Table \ref{tab:latency}. To balance performance, we provide a lightweight variant (denoted as Ours-S in Tables~\ref{tab:latency}, ~\ref{tab:supp_eval}): omit geometry guidance and apply scale guidance only at the penultimate denoising step. We further supplement a comprehensive evaluation of this lightweight version with current latest depth estimation methods in Table~\ref{tab:supp_eval}. Our system offers a favorable trade-off between accuracy and cost, supporting both efficient online alignment and more accurate but costlier offline optimization. 

\begin{table}[ht]
    \centering
    \scalebox{0.78}{
    \begin{tabular}{l|ccc}
        \toprule
        Method & Latency (s) & Max Memory (MiB) \\
        \midrule
        % ChronoDepth & 44.5 \\
        DepthCrafter \cite{hu2024depthcrafter}& 9.01 & 13449 \\
        DepthAnyVideo \cite{yang2024depth} & 14.6 & 24097 \\
        DUSt3R \cite{wang2024dust3r} & 430 & 9207 \\
        MonST3R \cite{zhang2024monst3r} & 572 & 30559 \\
        \midrule 
        Post Opt. & 652 & 14191 \\
        \midrule
        \textbf{Ours}: \\
        Scale Guidance Only  & 33.3 & 18223 \\
        Geometry Guidance Only & 952 & 26833\\
        Ours & 961 & 26833 \\
        Ours-S & 18.1 & 18223 \\
        \bottomrule
        \end{tabular}}
    \caption{\textbf{Inference Cost Evaluation} for 90 frames (Resolution: 640$\times$ 448) on a 40GB A100 GPU.}
\label{tab:latency}
\end{table}

\begin{table*}[h]
    \centering
    \scalebox{0.88}{
    \begin{tabular}{ll|cc|cc|cc|cc}
         \toprule
         \multicolumn{2}{c|}{Dataset} & \multicolumn{2}{c|}{ScanNet} & \multicolumn{2}{c|}{GMU Kitchen} & \multicolumn{2}{c}{KITTI} & \multicolumn{2}{c|}{Bonn}  \\
        \midrule
        \multicolumn{2}{c|}{Metrics} & AbsRel$\downarrow$ & $\delta_{1} \uparrow$  & AbsRel$\downarrow$ & $\delta_{1} \uparrow$ & AbsRel$\downarrow$ & $\delta_{1} \uparrow$  & AbsRel$\downarrow$ & $\delta_{1} \uparrow$ \\
        \midrule
        \multirow{7}{*}{\ding{172}} & ChronoDepth \cite{shao2024chrono} & 0.172 & 0.749 & 0.196 & 0.650 & 0.178 & 0.733 & 0.092 & 0.932  \\
        & DepthCrafter \cite{hu2024depthcrafter} & 0.141 & 0.799 & 0.143 & 0.795 & 0.114 & 0.879 & 0.095 & 0.917  \\
         % & DepthAnyVideo \cite{yang2024depth}\textsuperscript{*} &  {0.100} & {0.910} & {0.089} &  \underline{0.932} & 0.070 & 0.966 & {0.085} & {0.942} \\
        & DUSt3R \cite{wang2024dust3r}\textsuperscript{\dag} & \textbf{0.059} & \textbf{0.972} & \textbf{0.069} & \textbf{0.968} & 0.155 & 0.777 & 0.075 & 0.935  \\
        & MonST3R \cite{zhang2024monst3r}\textsuperscript{\dag} & \underline{0.067} & \underline{0.959} & \underline{0.087} & {0.920} & 0.192 & 0.692 & \textbf{0.046} & \textbf{0.975}  \\
        \cmidrule{2-10}
        & Ours-S & 0.128 & 0.834 & 0.126 & 0.844 & \underline{0.112} & \underline{0.882} & 0.070 & 0.968  \\
        & Ours & 0.113 & 0.870 & 0.113& 0.881 & \textbf{0.110} & \textbf{0.887} & \underline{0.069} & \underline{0.972} \\
        \midrule
        \multirow{7}{*}{\ding{173}} & ChronoDepth \cite{shao2024chrono} & 0.193 & 0.699 & 0.225 & 0.612 & 0.193 & 0.691 & 0.132 & 0.853  \\
        & DepthCrafter \cite{hu2024depthcrafter} & 0.171 & 0.716 & 0.160 & 0.748 & 0.173 & 0.727 & 0.143 & 0.801  \\
        % & DepthAnyVideo \cite{yang2024depth}\textsuperscript{*} & 0.127 & 0.840 & \textbf{0.113} & \textbf{0.883} & 0.103 & 0.911 & {0.094} & {0.922} \\
        & DUSt3R \cite{wang2024dust3r}\textsuperscript{\dag} & \textbf{0.090} & \textbf{0.934} &  \underline{0.127} & {0.828}  & 0.174 & 0.731 & 0.115 & 0.889 \\
        & MonST3R \cite{zhang2024monst3r}\textsuperscript{\dag} & \underline{0.099} & \underline{0.911} & 0.140 & 0.817 & 0.232 & 0.620 & \textbf{0.086} & {0.918}  \\
        \cmidrule{2-10}
        & Ours-S & 0.157 & 0.757 & 0.138 & 0.807 & \underline{0.119} & \underline{0.863} & 0.094 & \underline{0.926}  \\
        & Ours & 0.154 & 0.769 & 0.131 & \underline{0.837} & \textbf{0.117} & \textbf{0.869} & \underline{0.093} & \textbf{0.929}  \\
         \bottomrule
    \end{tabular}}
    \caption{\textbf{Supplementary Depth Evaluation}. Ours-S: a lightweight version of our method. \ding{172} and \ding{173}: the shortest and longest video length settings in our main paper. \textsuperscript{*}: trained on Hypersim. \textsuperscript{\dag}: trained on ScanNet++.}
\label{tab:supp_eval}
\end{table*}

\noindent \textbf{Cost Comparison with Post Optimization. } 
Our full system has significantly better results than post optimization with comparable cost, and our scale-only version is both faster (Table~\ref{tab:latency}) and superior (main paper Table 3) than post optimization. 

% \noindent \textbf{Comparison with Depth Any Video (DAV) \cite{yang2024depth}. } DAV supports at most 192 frames per video, so we compare with its original implementation only on our shortest video setting (Table~\ref{tab:supp_eval}~\ding{172}). For longer videos, we apply the same sliding window strategy as our method, running DAV on each segment and aligning the results. DAV demonstrates degraded performance on long sequences (\ding{173}) compared to shorter ones (\ding{172}) over four datasets. Our method outperforms DAV on Bonn and KITTI (Table~\ref{tab:supp_eval}), while DAV performs better on ScanNet and GMU Kitchen - likely due to its extensive training on indoor data (such as Hypersim). In contrast, our base model DepthCrafter has limited exposure to indoor scenes, which may constrain its performance in such environments.

\noindent \textbf{Comparison with DUSt3R \cite{wang2024dust3r} and MonST3R \cite{zhang2024monst3r} (3Rs).} We follow the depth estimation paradigm, taking a video as input and predicting video depths, while 3Rs use image pairs to predict point maps, adopting a fundamentally different technical route.
For comparison, we use sliding window strategy and align windows via overlap area to enable long video inference. We use 3Rs to adjacent frames in each window and derive depths from predicted point maps.  Results are shown in Table~\ref{tab:supp_eval} (DUSt3R fails on 2 of 26 Bonn scenes, so it's averaged over 24 scenes; others use all 26). While 3Rs perform better on indoor scenes, likely due to training on  ScanNet++, our method achieves better results on KITTI and Bonn.

% \section{Inference Latency Evaluation}
% \begin{table}[ht]
%     \centering
%     \scalebox{0.9}{
%     \begin{tabular}{l|cc}
%         \toprule
%         Method & Latency for 90 Frames (s) \\
%         \midrule
%         ChronoDepth & 44.5 \\
%         DepthCrafter & 9.01 \\
%         \midrule 
%         \textbf{Ours}: \\
%         Scale Guidance Only  & 33.3 \\
%         Geometry Guidance Only & 952 \\
%         Scale and Geometry Guidance & 961 \\
%         \bottomrule
%         \end{tabular}}
%     \caption{\textbf{Inference Latency Evaluation.}}
% \label{tab:latency}
% \end{table}

% We compare inference latency by evaluating 90-frame videos at 640 $\times$ 448 resolution on an A100 40G GPU with ChronoDepth and DepthCrafter (Table \ref{tab:latency}). For ChronoDepth, we use default settings from its official repository: chunk size = 5, n token = 10, denoising steps = 5, and sigma epsilon = -4.0. For DepthCrafter, we set denoising steps = 5 and window size = 90. For our method, we use a window size of 90 as well and apply guidance in the last two denoising steps.

% As shown in Table \ref{tab:latency}, scale guidance introduces minimal latency, even outperforming ChronoDepth in speed. As demonstrated in Table 1 and Table 3 in the main paper, scale guidance alone significantly improves cross window depth scale synchronization and depth accuracy on the Bonn dataset and ScanNet dataset. For cases requiring further consistency, geometry guidance can be optionally applied.

\section{Supplementary Ablation Study}
We present supplementary ablation studies on the guidance function, including the starting step, guidance order, and geometry guidance loss terms. These studies are conducted on the first 25 scenes of the ScanNet test set, with video length as 150, which is consistent with the ablation settings in the main paper.

\begin{table}[t!]
    \centering
    % \aboverulesep=0ex % Solution part 1 of 3
    % \belowrulesep=0ex
    \begin{tabular}{l|cc}
        \toprule
        Step & AbsRel & $\delta<1.25$ \\ 
        \midrule
        5 & 0.146 & 0.779 \\
        4 & 0.144 & 0.787 \\
        3 & 0.137 & \textbf{0.803} \\
        2 & \textbf{0.137} & 0.801 \\
        1 & 0.151 & 0.757 \\
        \midrule
        Order & AbsRel & $\delta<1.25$ \\
        \midrule
        Scale First & 0.145 & 0.782 \\
        Geometry First & \textbf{0.137} & {0.801} \\
        \bottomrule    
    \end{tabular}
    \caption{\textbf{Supplementary Ablation Study on Guidance Implementations. } The "Step" column indicates the number of final steps in the diffusion  loop at which the guidance is employed. For example, if the "Step" value is 5, it means that the guidance starts from the last five steps.}
\label{tab:start}
\end{table}

\noindent\textbf{Starting Step of Guidance.}
We conduct experiments on when to start guidance. As shown in Table~\ref{tab:start}, starting guidance early does not always improve performance, as noise in the predicted clean latent during early denoising stages can mislead the guidance process. Starting from the third-to-last step yields the highest depth accuracy but the same relative error with second-to-last and adds 50\% inference latency. Thus, starting from the second-to-last step strikes the best balance between efficiency and effectiveness.

\noindent\textbf{Ablation on Guidance Term Order.} 
As shown in Table~\ref{tab:start}, applying the geometry guidance first and then the scale guidance leads to better depth estimation results, indicating that a geometric aligned intra-window depth help a better synchronization of depth scale between windows.

\noindent\textbf{Ablation on Each Loss Term in Geometry Guidance.}
We present ablation results on the effectiveness of each loss term in geometry guidance in Table \ref{tab:term}. Applying global geometric constraints (depth reprojection loss and tracking loss) individually yield greater improvements than local constraints (surface normal loss and smoothness loss). Depth reprojection loss outperforms tracking loss, as it operates on every pixel rather than sparse tracked points. Combining local constraints with global constraints further enhances the performance, as global constraints alone just align the depths between frames but may not converge to the optimal solution, while local constraints introduces supplementary information about the geometric structure to the optimization process and help avoid local optima.

\begin{table}[t!]
    \centering
    % \aboverulesep=0ex % Solution part 1 of 3
    % \belowrulesep=0ex
    \begin{tabular}{c|cc}
        \toprule
        \multirow{2}{*}{Constraints} & \multicolumn{2}{c}{Metrics} \\
        \cmidrule{2-3}
        & AbsRel & $\delta<1.25$ \\ 
        \midrule
        Baseline 90 & 0.136 & 0.816 \\
        loss d only & 0.122& 0.856 \\
        loss n only & 0.121 & 0.851 \\
        loss s only & 0.122 & 0.847 \\
        loss t only & 0.121 & 0.850 \\
        \midrule
        all loss & \textbf{0.108} & \textbf{0.880} \\
        \bottomrule    
    \end{tabular}
    \caption{\textbf{Ablation Study on Guidance Terms} on ScanNet first 25 Test Scenes.}
\label{tab:term}
\end{table}

\section{Supplementary Qualitative Examples}
\begin{figure*}[h]
    \centering
    \includegraphics[width=\textwidth]{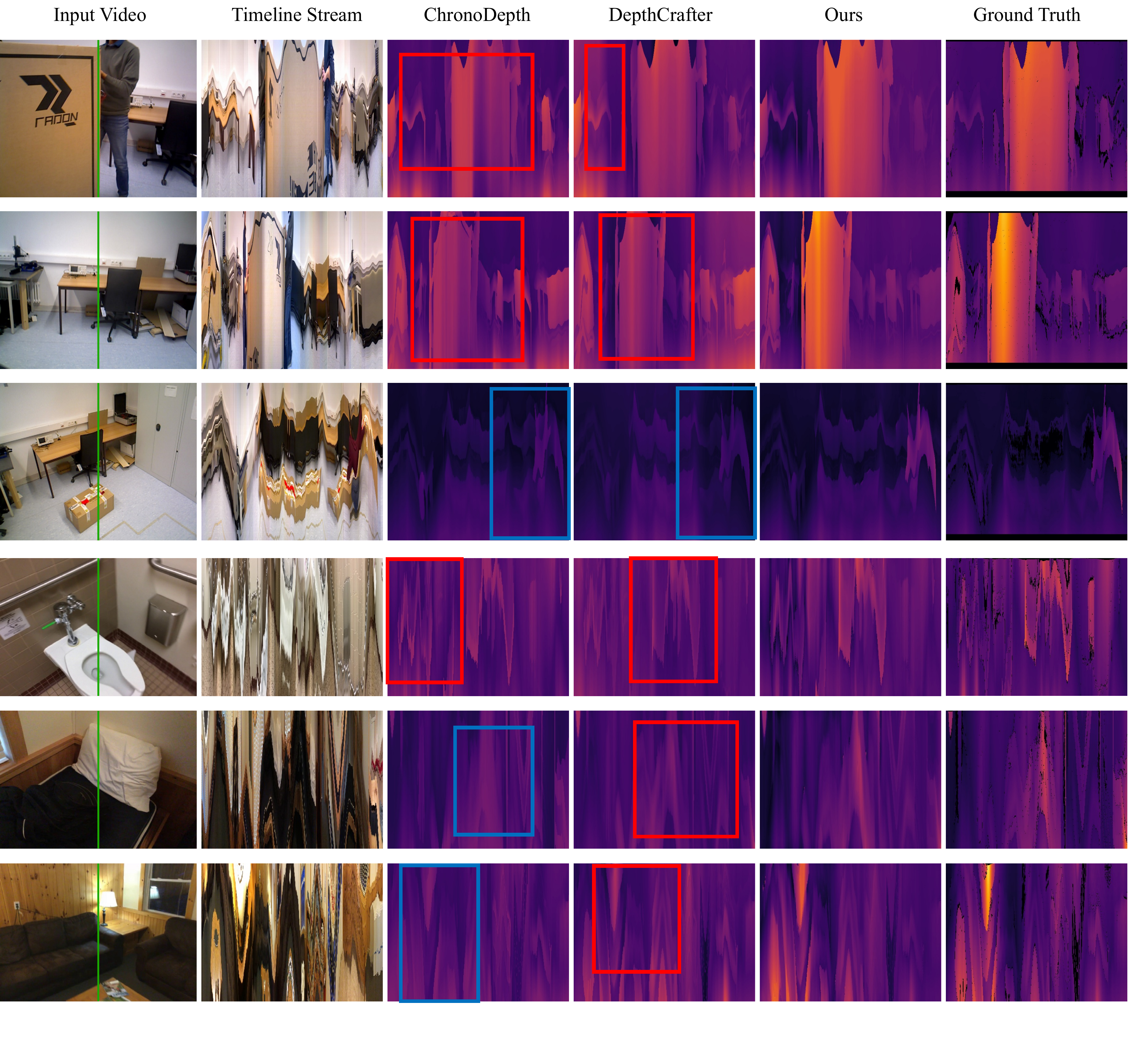}
    \caption{\textbf{Supplementary Qualitative Comparison on Long Video Depth Consistency.} We compare predictions on 450-frame ScanNet videos and 590-frame Bonn videos. Slicing along the timeline at the green-line position (first column), we concatenate results to visualize temporal changes in color and depth. Red boxes highlight depth scale inconsistencies, while blue boxes mark depth inaccuracies in previous methods. Our method shows superior depth accuracy and scale consistency over time.
    }
    \label{fig:timeseq_supp}
    % \vspace{-0.5cm}
\end{figure*}

\begin{figure*}[h]
    \centering
    \includegraphics[width=0.8\textwidth]{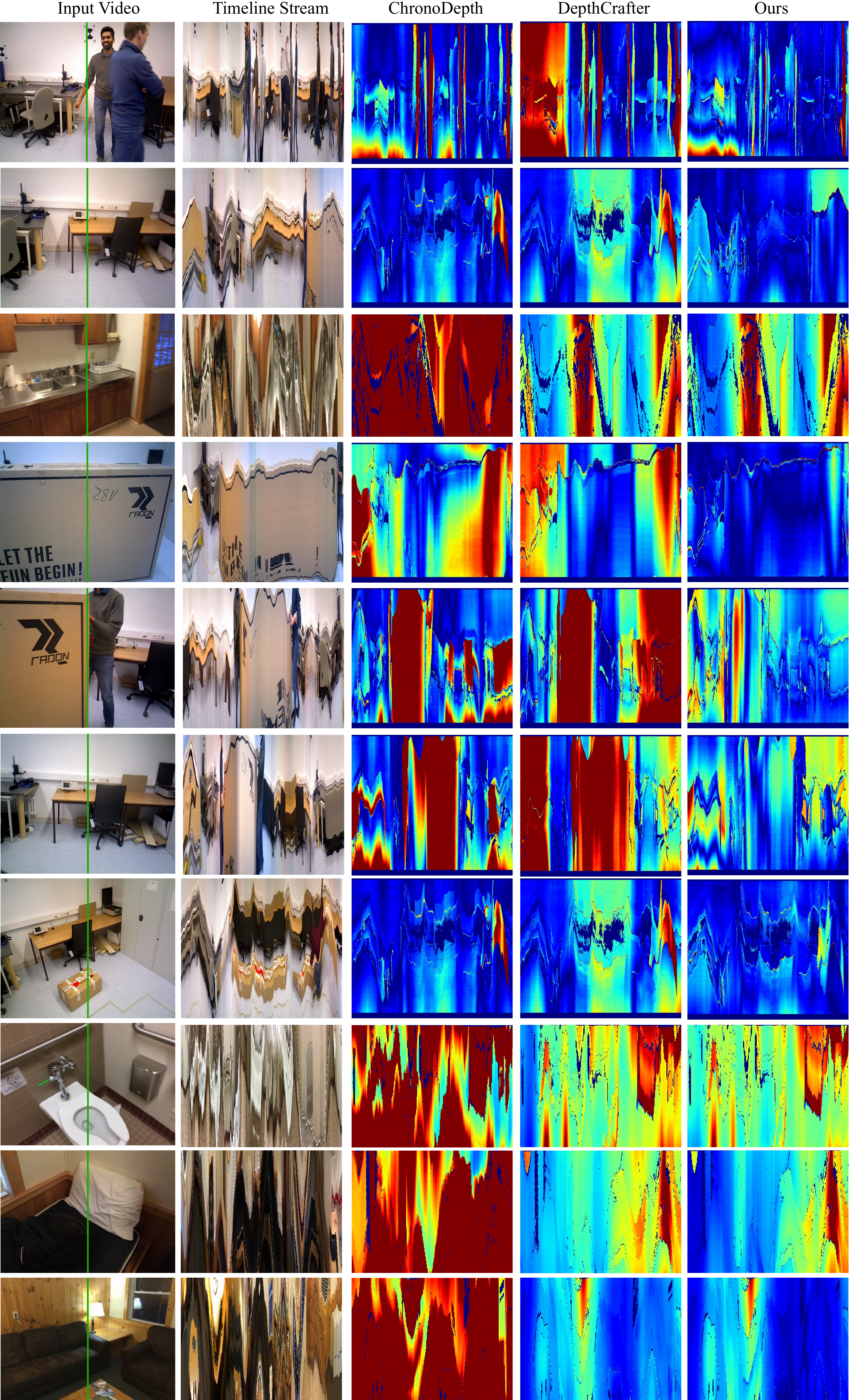}
    \caption{\textbf{Depth Error Map Comparison on Long Video Depth Consistency.} Supplementary visualization of absolute error maps for qualitative analysis (see Figure 4 in the main paper and Figure~\ref{fig:timeseq_supp}). Blue indicates low error, while red corresponds to high error.}
    \label{fig:timeseq_supperr}
    % \vspace{-0.5cm}
\end{figure*}

\begin{figure*}[t!]
    \centering
    \includegraphics[width=\textwidth]{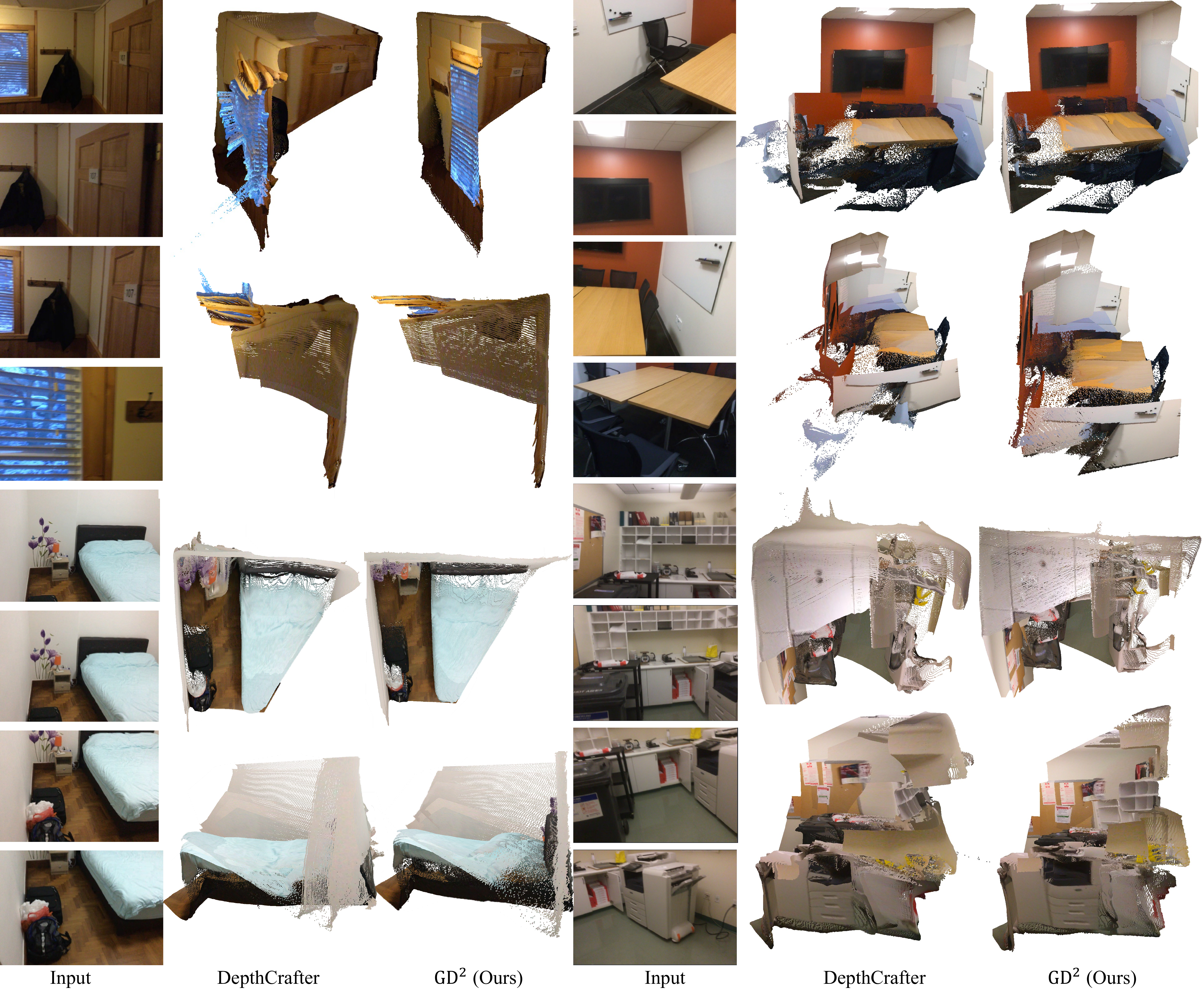}
        \caption{\textbf{Supplementary Qualitative Examples for 3D Reconstruction Comparisons}. } 
    \label{fig:q1}
    % \vspace{-0.5cm}
\end{figure*}

\begin{figure*}[t!]
    \centering
    \includegraphics[width=\textwidth]{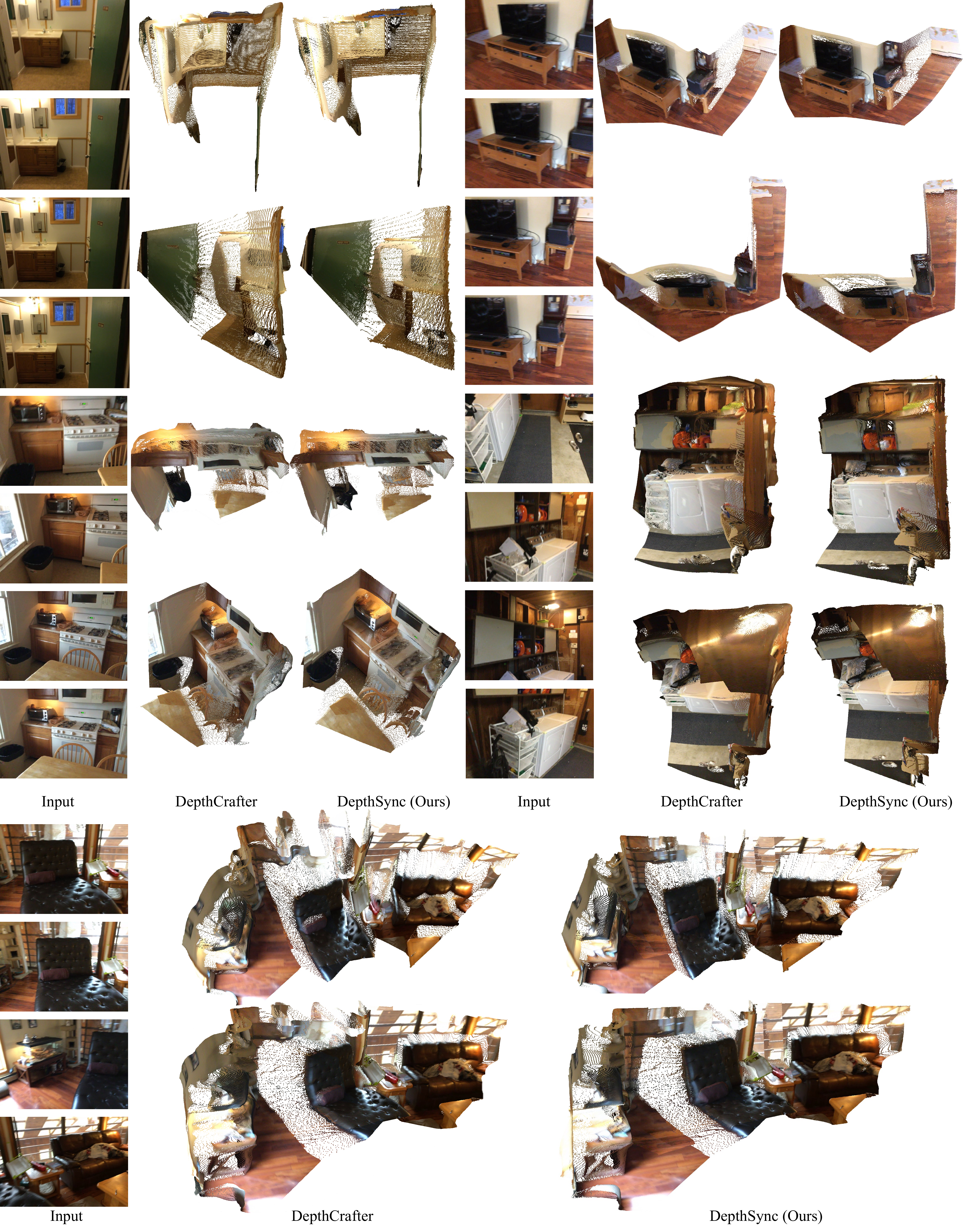}
    \caption{\textbf{Supplementary Qualitative Examples for 3D Reconstruction Comparisons}. } 
    \label{fig:q2}
    % \vspace{-0.5cm}
\end{figure*}

% In Figures \ref{fig:q1}, \ref{fig:q2}, \ref{fig:q4} and \ref{fig:q5}, 
We provide additional qualitative examples to supplement the main paper.  Figure \ref{fig:timeseq_supp}, Figure \ref{fig:timeseq_supperr}, Figure \ref{fig:q1} and Figure \ref{fig:q2} demonstrate that our DepthSync achieves better scale synchronization between windows for long videos and more accurate geometric consistency within each window.


% \begin{figure*}[t!]
%     \centering
%     \includegraphics[width=\textwidth]{fig/supp_qualitative2.pdf}
%         \caption{\textbf{Supplementary Qualitative Examples for 3D Reconstruction Comparisons}. } 
%     \label{fig:q3}
%     % \vspace{-0.5cm}
% \end{figure*}

\end{document}